\documentclass{article}

    \usepackage[final]{neurips_2024}

\usepackage[utf8]{inputenc} 
\usepackage[T1]{fontenc}    
\usepackage{hyperref}       
\usepackage{url}            
\usepackage{booktabs}       
\usepackage{amsfonts}       
\usepackage{nicefrac}       
\usepackage{placeins}       

\usepackage{microtype}      
\usepackage{xcolor}         
\usepackage{graphicx}
\usepackage{caption}
\usepackage{enumitem}
\usepackage{caption}
\usepackage{subcaption}
\usepackage{courier}
\usepackage{booktabs}
\usepackage{amsmath}
\usepackage{array}
\usepackage{listings}
\usepackage{xcolor}

\definecolor{codebackground}{rgb}{0.95,0.95,0.95}
\definecolor{codestring}{rgb}{0.6,0.1,0.1}
\definecolor{codekeyword}{rgb}{0.1,0.1,0.6}
\definecolor{codecomment}{rgb}{0.2,0.6,0.2}

\lstdefinestyle{mystyle}{
    backgroundcolor=\color{codebackground},
    commentstyle=\color{codecomment},
    keywordstyle=\color{codekeyword},
    stringstyle=\color{codestring},
    basicstyle=\ttfamily\footnotesize,
    breakatwhitespace=false,
    breaklines=true,
    captionpos=b,
    keepspaces=true,
    showspaces=false,
    showstringspaces=false,
    showtabs=false,
    tabsize=2
}

\title{Evaluating Sparse Autoencoders\\on Targeted Concept Erasure Tasks}

\author{
Adam Karvonen\thanks{Equal contribution. Correspondence to  \href{mailto:adam.karvonen@gmail.com}{adam.karvonen@gmail.com} and \href{mailto:can.rager@gmail.com}{can.rager@gmail.com}.} \\Independent \And 
Can Rager$^{*}$\\Independent\\ \And
Samuel Marks\\Anthropic \And
Neel Nanda
}

\begin{document}

\maketitle


\begin{abstract}
Sparse Autoencoders (SAEs) are an interpretability technique aimed at decomposing neural network activations into interpretable units. However, a major bottleneck for SAE development has been the lack of high-quality performance metrics, with prior work largely relying on unsupervised proxies. In this work, we introduce a family of evaluations based on SHIFT, a downstream task from Marks et al.~\cite{marks2024sparsefeaturecircuitsdiscovering} in which spurious cues are removed from a classifier by ablating SAE features judged to be task-irrelevant by a human annotator. We adapt SHIFT into an automated metric of SAE quality; this involves replacing the human annotator with an LLM. Additionally, we introduce the Targeted Probe Perturbation (TPP) metric that quantifies an SAE's ability to disentangle similar concepts, effectively scaling SHIFT to a wider range of datasets. We apply both SHIFT and TPP to multiple open-source models, demonstrating that these metrics effectively differentiate between various SAE training hyperparameters and architectures. 
\end{abstract}

\section{Introduction}


Sparse autoencoders (SAEs) have emerged as a promising tool for neural network interpretability. This year alone, a broad range of SAE architectures and training approaches has been proposed, including TopK \cite{gao2024scalingevaluatingsparseautoencoders}, Gated \cite{rajamanoharan2024improvingdictionarylearninggated}, BatchTopK \cite{bussmann2024batchtopk}, p-Annealing \cite{karvonen2024measuringprogressdictionarylearning}, and JumpReLU SAEs \cite{rajamanoharan2024jumpingaheadimprovingreconstruction}. However, the field of dictionary learning for neural network interpretability faces a fundamental challenge: the lack of trusted metrics to evaluate progress. Unlike traditional machine learning tasks with clear objective functions, SAE development operates without a well-defined "ground truth" for interpretability. 

The unsupervised proxy metrics \textit{sparsity} and \textit{fidelity} commonly used in prior work \cite{cunningham2023sparseautoencodershighlyinterpretable, lieberum2024gemmascopeopensparse, marks2024sparsefeaturecircuitsdiscovering} do not always correlate with the desirable characteristics we seek, such as interpretability \cite{jermyn2024tanh, braun2024identifyingfunctionallyimportantfeatures}. This misalignment between optimization targets and interpretability goals creates a significant hurdle in assessing whether proposed improvements genuinely advance the field or merely optimize for proxy objectives.
While SAEs are intended to faithfully decompose model activations into interpretable units, our limited understanding of model internals makes it difficult to establish a definitive benchmark. To address this, we propose directly measuring SAEs through their applicability to downstream tasks.

Sparse Human Interpretable Feature Trimming (SHIFT) is a method from Marks et al.~\cite{marks2024sparsefeaturecircuitsdiscovering} that debiases a classifier by removing spurious correlates in models. This method allows a researcher to inspect the concepts that causally influence a neural network classifier's outputs, and selectively remove concepts which seem irrelevant to the intended classification task. We introduce a family of evaluations based on SHIFT called Spurious Correlation Removal (SCR) which measures an SAE's ability to disentangle concepts and remove spurious correlations.

However, the reliance of SCR on datasets containing potential spurious correlations makes it challenging to scale to a wider variety of concepts. Motivated by this, we additionally develop Targeted Probe Perturbation (TPP), a generalization of SCR to all text classification datasets. Given a dataset, TPP evaluates an SAE's quality by measuring its ability to identify and modify one specific the class while leaving other classes unchanged. We specifically select a small number of related classes, train linear probes for each, delete the SAE latents most causally relevant to one probe, and measure the degradation of that probe relative to the rest. For both SCR and TPP, we select SAE latents using probe attribution scores, and optionally filter the latents for interpretability using an LLM judge. We evaluate our evaluation metrics against a range of sanity checks, looking at the improvement throughout training and unimodality in sparsity. 

Our main contributions are as follows: We adapt the spurious correlation removal task in SHIFT \cite{marks2024sparsefeaturecircuitsdiscovering} as an SAE evaluation metric and generalize this technique to text classification datasets by introducing the Targeted Probe Perturbation (TPP) metric.
Second, we train and open-source a suite of SAEs and evaluate our metrics across multiple language models, datasets, and SAE training checkpoints.

\section{Background}
\label{sec:background}

\textbf{Sparse autoencoders (SAEs)} aim to identify an overcomplete basis of sparse, interpretable features from model internal representations \cite{sharkey2022interim, bricken2023monosemanticity, elhage2022superposition}. The quality of SAEs is determined by their faithfulness to the model's internal computations and their ability to disentangle human-interpretable concepts. This disentanglement can be further broken down into correlational and causal aspects, namely the detection of interpretable concepts and the ability to modify model behavior in targeted ways.

Unsupervised metrics are the current standard for evaluating SAEs \cite{rajamanoharan2024improvingdictionarylearninggated, gemmateam2024gemma2improvingopen, cunningham2023sparseautoencodershighlyinterpretable}: (1) the cross-entropy loss recovered, which measures how well the original model's loss can be reconstructed using the SAE's predictions, and (2) the L0-norm of feature activations, which quantifies the sparsity of activated features for a given input \cite{ferrando2024primerinnerworkingstransformerbased}. Recent work has explored evaluating SAEs on board game models \cite{karvonen2024measuringprogressdictionarylearning}, features in manually identified circuits \cite{makelov2024subspace} and detecting pre-defined natural language concepts \cite{gao2024scalingevaluatingsparseautoencoders}.

\textbf{Concept removal} aims to identify and eliminate specific concepts or biases from neural representations while preserving overall performance on downstream tasks. Concept erasure in deep linear classifiers was pioneered by Bolukbasi et al. \cite{bolukbasi2016man} by using PCA. Ravfogel et al. \cite{ravfogel-etal-2020-null} iteratively trained linear classifiers to identify and remove undesired concepts. Multiple techniques have been proposed since then, ranging from linear methods like Hard Debias \cite{wang-etal-2020-double}, R-LACE \cite{pmlr-v162-ravfogel22a} and LEACE \cite{belrose2023leace} to more complex non-linear techniques \cite{iskander-etal-2023-shielded, ravfogel-etal-2022-adversarial}. In this work, our goal is not to improve on the state of the art for concept erasure, but rather to adapt concept erasure tasks into SAE progress metrics.

\section{Method}
\label{sec:method}

In this paper, we focus on the causal isolation of concepts as our primary metric for SAE quality. Our approach quantifies how effectively an SAE can manipulate individual concepts within the model's representations. To illustrate our methods, consider the task of selecting the best SAE from a suite of SAEs. Given a list $\mathcal{C}$ of concepts in natural language, we follow three steps to systematically evaluate each SAE in the suite:

\begin{enumerate}
    \item For each concept $c\in\mathcal{C}$, train a binary classification head on our model for $c$.
    \item Identify a set $\mathcal{L}_c$ of SAE latents corresponding to each concept $c \in\mathcal{C}$.
    \item Measure whether ablating features corresponding to concept c have the expected effect on the classifiers from (1).    
\end{enumerate}


Intuitively, an SAE is high quality if for each concept $c\in\mathcal{C}$ ablating the features $\mathcal{F}_c$ associated to c has a large affect on our classifier's accuracy for c, but not on our classifiers for other concepts. Our SHIFT and TPP metrics are different ways of operationalizing and quantifying this intuition.

  To validate our metrics, we conduct a series of sanity checks. These checks verify that our metric aligns with fundamental properties of SAEs, such as their characteristic sparsity and their expected performance improvements over the course of training. The subsections below describe each step in the evaluation process in detail. Appendix \ref{apx:sae} contains further details about our SAE training process.

\subsection{SAE Latent Selection}
\label{sec:method:latent_selection}

Measuring the disentanglement of a given natural language concept with an SAE, requires identifying the subset of corresponding SAE latents. We determine relevant SAE latents by ranking their causal effect on a concept-related probe and optionally filter for interpretability as judged by an LLM.

\textbf{Probing.} The classification heads are realized by training binary linear probes to detect a concept $c$ from model activations as described in Appendix \ref{apx:probe}. We identify the direct effect of individual SAE latents to classification heads via attribution patching \cite{nanda2023attribution}. The classification head is directly applied to the output of the SAE in our setup. Hence the direct effect of a latent to the classification head is given by the dot product. The attribution score of latent with index $a$ on concept $c$ is given by

\begin{equation}
I(\mathbf{a}, c) = (\mathbf{d}_a \cdot \mathbf{P})(\mathbf{a}_\text{pos} - \mathbf{a}_\text{neg}) ,
\end{equation}

where $\mathbf{a}$ denotes the batch of activations of latent $a$, $\mathbf{d}_a$ denotes the SAE decoder vector corresponding to $a$, $\mathbf{P}$ is the weight matrix of the binary probe, $\mathbf{a}_\text{pos}$ is the mean activation of the latent for inputs of the targeted concept, and $\mathbf{a}_\text{neg}$ is the mean activation for inputs unrelated to the concept.
The difference of mean activations ($\mathbf{a}_\text{pos} - \mathbf{a}_\text{neg}$) helps filter out high-frequency latents that activate with equal frequency on positive and negative class inputs, focusing on latents that discriminate between the classes.

We select the top $N$ latents with the highest attribution scores\footnote{The computation of attribution scores $I$ can be efficiently calculated using precomputed model activations. However, this is restricted to the SAE layer coinciding with the probe layer. If probe layer and SAE layer differ, the attribution $I$ can be approximated with attribution patching \cite{nanda2023attribution}, which requires requires additional forward and backward passes on the order of \texttt{num\_classes * num\_batches}.}.
Our current approach involves increasing $N$ until we observe statistically significant differences in the performance of various SAEs. We find that the required $N$ increased with model size. However, the optimal selection of $N$ is an open problem, as there isn't a clear optimal number of latents that a good SAE should have for a given concept.

\textbf{Automated intepretability judgement.} We optionally employ Claude-3.5 Sonnet as an LLM judge to decide whether an SAE latent is interpretable. The LLM judge receives a description similar to EleutherAI \cite{juang2024opensourceautointerp}: Our prompt contains the top 5 contexts that activated the latent from a set of 10k random web text samples \cite{gao2020pile800gbdatasetdiverse}, as well as the top 5 promoted tokens indicated by direct logit attribution \cite{nostalgebraist2020interpreting}, and 4 few-shot examples demonstrate scoring the relation to each concept from 0 to 4. The model is instructed to perform in-context reasoning and outputs a score for each category in json format.


Appendix \ref{apx:prompt} shows a full example prompt. We observe that the LLM judge is prone to false negatives, rating concepts with 0 although they are related. We decide to keep latents that show any relation with score 1 or higher. We refined the system prompts by manually labelling 60 examples, and looking at prompts where ratings differ. After tweaking the prompts the inter-rater agreement shows a Cohen's $\kappa$ value of 0.44 for gender and 0.58 for profession, passing our minimum requirements.\footnote{For reference, $\kappa=0$ means random chance, $\kappa=1$ means perfect correlation and $\kappa > 0.4$ denotes reasonable correlation.}.

\subsection{SHIFT}
\label{sec:method:shift}

 In the SHIFT method \cite{marks2024sparsefeaturecircuitsdiscovering}, a human evaluator debiases a classifier by ablating SAE latents. We automate SHIFT and use it as an evaluation metric for SAE quality. Better SAEs enable a more precise removal of spurious correlations, thereby effectively debiasing the classifier.

The SHIFT method requires a dataset that maps text to two binary labels. Here, we use the Bias in Bios dataset \cite{De_Arteaga_2019}, which maps professional biographies to occupation and gender, and the Amazon reviews dataset \cite{hou2024bridging}, which covers reviews from a broad range of product categories with accompanying rating scores.
First, we filter both datasets for two binary labels. For example, from the Bias in Bios dataset, we select two professions ({professor}, {nurse}) and the gender labels ({male}, {female}). We partition this dataset into a balanced set—containing all combinations of {professor}/{nurse} and {male}/{female}—and a biased set that only includes {male+professor} and {female+nurse} combinations. We then train a linear classifier $C_\text{b}$ on the biased dataset, which we will attempt to debias.


We perform latent selection as described in section \ref{sec:method:latent_selection} which replicates the original SHIFT setup by Marks et al. \cite{marks2024sparsefeaturecircuitsdiscovering}. In it, we select the top $n$ SAE latents $\mathcal{L}$ according to their \textit{absolute} probe attribution score to the biased probe. We then filter $\mathcal{L}$ using an LLM judge, retaining only the latents that the LLM scores as \textit{unrelated} to our desired concept. 

Alternatively, we define a spurious-feature-informed method to identify the SAE latents without requiring an LLM judge: We select set $L$ containing the top $n$ SAE latents according to their absolute probe attribution score with a probe trained specifically to predict the spurious signal.
\footnote{Our spurious-feature-informed latent selection corresponds to the "feature skyline" in Marks et al. \cite{marks2024sparsefeaturecircuitsdiscovering}.}

For each original and spurious-feature-informed set $\mathcal{L}$ of selected features, we remove the spurious signal by defining a modified classifier $C_\text{m} = C_\text{b} \backslash\ L $ where all selected \textit{unrelated yet with high attribution} latents are zero-ablated. The accuracy with which the modified classifier $C_\text{m}$ predicts the desired class when evaluated on the balanced dataset indicates SAE quality. A higher accuracy suggests that the SAE was more effective in isolating and removing the spurious correlation of e.g. gender, allowing the classifier to focus on the intended task of e.g. profession classification. We consider a normalized evaluation score 
\begin{equation}
S_{\text{SHIFT}} = \frac{A_{\text{abl}} - A_{\text{base}}}{A_{\text{oracle}} - A_{\text{base}}}
\label{eq:s_shift}
\end{equation}
where $A_{\text{abl}}$ is the probe accuracy after ablation, $A_{\text{base}}$ is the baseline accuracy (spurious probe before ablation), and $A_{\text{oracle}}$ is the skyline accuracy (probe trained directly on the desired concept). This score represents the proportion of improvement achieved through ablation relative to the maximum possible improvement, allowing fair comparison across different classes and models.

\subsection{Targeted Probe Perturbation}
\label{sec:method:tpp}

SHIFT requires datasets with correlated labels. We generalize SHIFT to all multiclass NLP datasets by introducing the targeted probe perturbation (TPP) metric. On a high level, we aim to find sets of SAE latents that disentangle the dataset classes. Inspired by SHIFT, we train probes on the model activations and measure the effect of ablating sets of latents on the probe accuracy. Ablating a disentangled set of latents should only have an isolated causal effect on one class probe, while leaving other class probes unaffected.


We consider a dataset mapping text to exactly one of $m$ concepts $c \in \mathcal{C}$. For each class with index $i = 1, \ldots, m$ we select the set $\mathcal{L}_i$ of the most relevant SAE latents as described in section \ref{sec:method:latent_selection}. Note, that we select the top \textit{signed} importance scores, as we are only interested in latents that actively contribute to the targeted class.

For each concept $c_i$, we partition the dataset into samples of the targeted concept and a random mix of all other labels.
We define the model with probe corresponding to concept $c_j$ with $j = 1, \ldots, m$ as a linear classifier $C_j$ which is able to classify concept $c_j$ with accuracy $A_j$. Futher, $C_{i, j}$ denotes a classifier for $c_j$ where latents $\mathcal{L}_i$ are ablated. Then, we iteratively evaluate the accuracy $A_{i, j}$ of all linear classifiers $C_{i, j}$ on the dataset partitioned for the corresponding class $c_j$. The targeted prope perturbation score

\vspace{-0.3cm}
\begin{equation}
    S_\text{TPP} = 
    \underset{(i = j)}{\text{mean}} ( A_{i, j} -A_{j} ) - \underset{(i \neq j)}{\text{mean}} ( A_{i, j} -A_{j} )
\label{eq:s_tpp}
\end{equation}

represents the effectiveness of causally isolating a single probe. Ablating a disentangled set of latents should only show a significant accuracy decrease if $i = j$, namely if the latents selected for class $i$ are ablated in the classifier of the same class $i$, and remain constant if $i \neq j$.


\section{Results}
\label{sec:results}

We trained and open-sourced a suite of SAEs (detailed in Appendix \ref{apx:sae}) to compare the introduced metrics SCR and TPP with the commonly used measure of sparsity. Since the training of SAEs require significant fractions of the model's training data to identify nuanced, interpretable features \cite{templeton2024scaling, gao2024scalingevaluatingsparseautoencoders} we evaluate SAEs at multiple checkpoints throughout training.

\subsection{SHIFT}
\label{sec:results:shift}

First, we evaluate both Pythia-70M and Gemma-2-2B SAEs on the SHIFT task. SAEs effectively remove an unwanted signal and improve the accuracy of a biased classifier for both models. The evaluation clearly differentiates SAEs of different sparsities and architectures in all experiments. 

\begin{figure}[htb!]
\centering
\begin{tabular}{cc}
\includegraphics[width=0.49\textwidth]{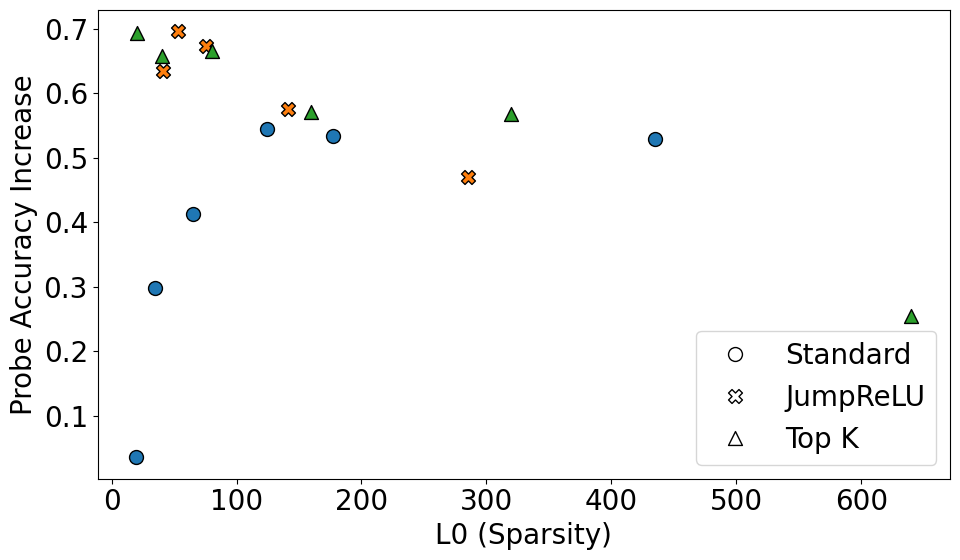} &
\includegraphics[width=0.49\textwidth]{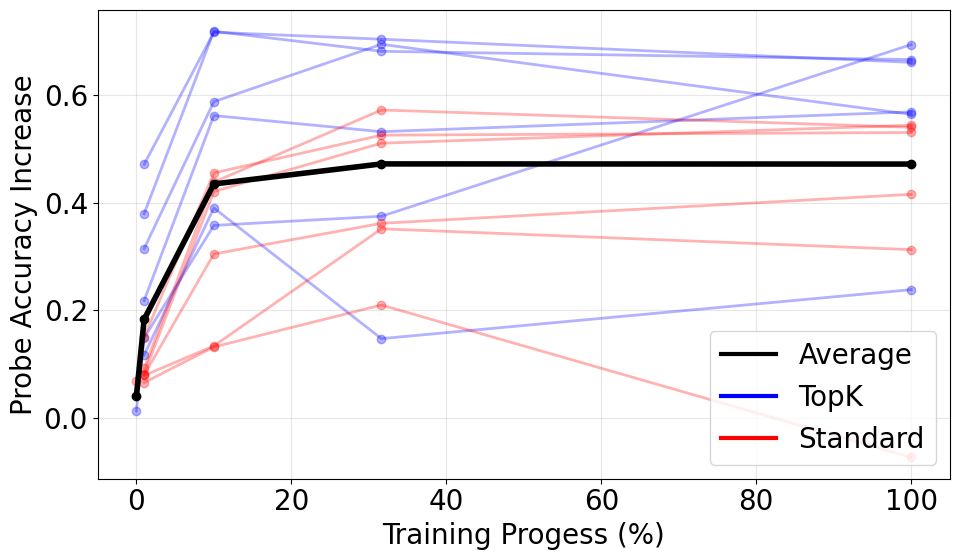}
\end{tabular}
\caption{The left scatterplot of loss recovered vs L0, with color corresponding to coverage score, and each point representing a single SAE. We differentiate between SAE training methods with shapes (left) and colors (right).}
\label{fig:gemma_scr_plots}
\end{figure}

Figure \ref{fig:gemma_scr_plots} shows the SHIFT evaluation of Gemma-2-2B SAEs trained on layer 19.
The metric reflects the relative accuracy increase $S_\text{SHIFT}$
of classifying the desired attribute after ablating spurious SAE latents as given in Equation \ref{eq:s_shift}. Here, we select the top 50 spurious latents from the original SHIFT method (Section \ref{sec:method:shift}) and filtered for interpretablity with an LLM judge (Section \ref{sec:method:latent_selection}).
Figure \ref{fig:gemma_scr_plots} (left) demonstrates a clear separation between architectures, with TopK and JumpReLU  outperforming the Standard architecture. Both TopK and JumpReLU are unimodal with respect to sparsity and peak in the L0 range [20, 100]. 
Figure \ref{fig:gemma_scr_plots} (right) verifies that SAEs improve on the SHIFT metric throughout training on average. The first 10\% of training (corresponding to 20M tokens) correspond to 85 \% of SHIFT score on average.


%

Our findings replicate across models, datasets, and latent selection methods, as further discussed in Appendix \ref{apx:amazon_scr}. Interestingly, most SAE latents identified by the spurious-feature-informed selection method are judged interpretable by the LLM. This trend especially holds for the peak region of the L0 range [20, 100].
The noise-informed method, which does not rely on an LLM judge, is very fast and a good proxy for the original SHIFT method.

In the original experiment by Marks et al. \cite{marks2024sparsefeaturecircuitsdiscovering}, SHIFT relied on Standard SAEs at many locations in the model and attribution patching. We find that with improved architectures such as TopK, good performance can be obtained with a single SAE. Moreover, we find that SHIFT scores before and after applying the LLM interpretability filter correlate highly with $r = 0.81$.  Appendix \ref{apx:probe} shows a direct comparison for Pythia-70M.

\subsection{TPP}
\label{sec:results:tpp}

\begin{figure}[htb!]
\centering
\begin{tabular}{cc}
\includegraphics[width=0.49\textwidth]{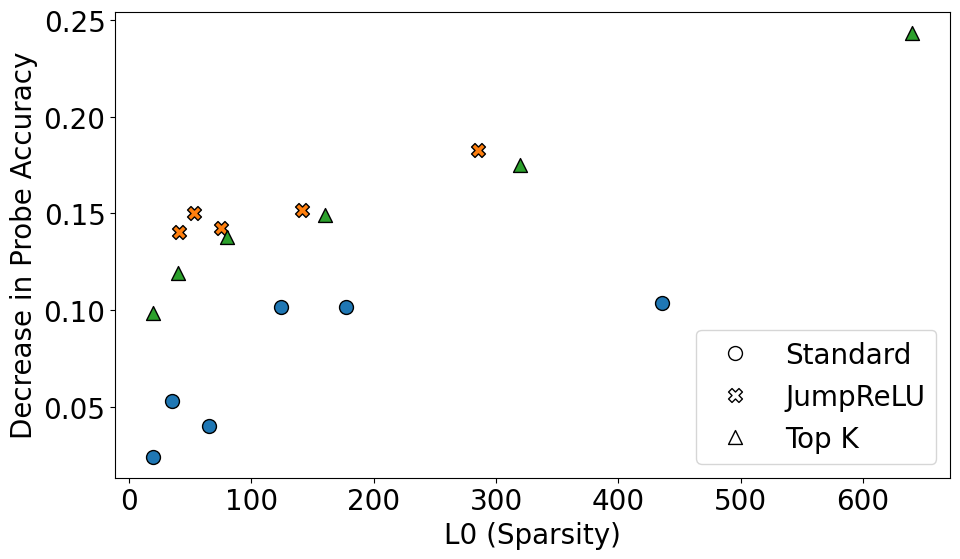} &
\includegraphics[width=0.49\textwidth]{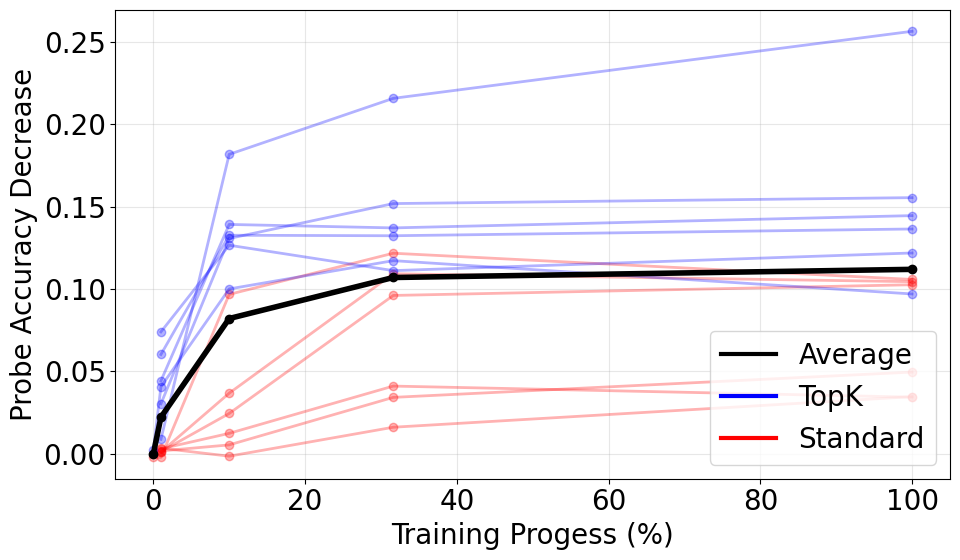}
\end{tabular}
\caption{ Targeted Probe Perturbation (TPP) scores over sparsity for SAEs of Standard, JumpReLU, and TopK architectures (left). TPP scores as a function of training progress, measured for checkpoints at 0\%, 1\%, 10\%, 31\%, and 100\% of SAE training over 6 TopK and 6 Standard SAEs (right). Each datapoint (left) and line (right) corresponds to a single SAE, architectures are differentiated by color.
}\label{fig:gemma_tpp_plots}
\end{figure}

Figure \ref{fig:gemma_tpp_plots} shows the TPP score (Equation \ref{eq:s_tpp}) of Gemma-2-2B SAEs trained on layer 19. Again we ablate up to 50 SAE latents. TPP clearly differentiates TopK and JumpReLU SAEs from the Standard SAEs in the left subplot. This separation is also visible in the TPP score evolution throughout training (right): While TopK and JumpReLU SAEs achieve 80 \% of their final TPP score after 10\% of training, Standard SAEs achieve 80 \% of only after 31\% of training. On average, the TPP score increases throughout training.
Moreover, we find that the TPP scores before and after applying the LLM interpretability filter correlate highly with $r = 0.88$. 
Appendix \ref{apx:pythia} contains results of TPP on Pythia-70M in Figure \ref{fig:pythia_plots}.

\section{Discussion and Limitations}

Our experimental results demonstrate that SHIFT and TPP effectively differentiate Standard SAEs from TopK and JumpReLU SAEs. While both metrics identify at least one SAE that successfully solves the task at hand, the optimal sparsity for each metric differs significantly. Further investigation on correlations of the TPP metric with L0 is required.


Our LLM judge has a lower bar for interpretability than the common implementation by Juang et al. \cite{juang2024opensourceautointerp}. While we simply score a latent's relevance to certain concepts, automated interpretability usually aims to find a concise description in natural language which is predictive of latent activations. 

SHIFT and TPP without the LLM judge are simpler, faster, and cheaper, and can be computed in seconds, enabling frequent evaluations such as during SAE training. However, we can get increased confidence in the metric using an LLM judge to ensure that identified latents are interpretable. Thus, there is a tradeoff in adding the LLM judge. We analyze the correlation of the metrics with and without the LLM judge in Appendix \ref{apx:judge_corr}. 

Our evaluation metrics rely on subjective hypotheses about what SAEs should learn based on human-understandable concepts, which may not accurately reflect the model's true internal representations. This dependency on human-generated concepts can overlook important latents and constrains the evaluation's scope.





\section{Conclusion}



SHIFT and our Targeted Probe Perturbation (TPP) method offer several advantages for evaluating Sparse Autoencoders (SAEs). They can be easily applied to a wide range of datasets, show improvement throughout training, and are computationally efficient. Both metrics can be computed in seconds when using precomputed activations, and we encourage researchers to use our codebase to evaluate their SAEs and monitor their SAE training runs.



However, SHIFT and TPP have some limitations, including complexity (especially when using an LLM judge) and undetermined hyperparameters. Thus, they should only be treated as additional evidence as part of a broader SAE evaluation suite. The lack of ground truth features and significant variability across model sizes makes hyperparameter and SAE latent selection challenging. Therefore, it is important to perform sanity checks on SAE evaluations, like monitoring metrics during training and evaluating SAEs across a range of sparsities. 

Our metrics focus on the causal isolation of human-interpretable concepts. However, high-quality SAE latents should also exhibit characteristics such as sparsity, disentanglement of natural language concepts, human interpretability, and correlation with natural language concepts. Developing evaluations that cover all these aspects of SAE quality remains an open challenge.

\clearpage
\section*{Acknowledgments}
This work was conducted as part of the ML Alignment \& Theory Scholars (MATS) Program. We want to express our sincere gratitude to McKenna Fitzgerald for her guidance and support during the
program. We want to extend our gratitude to Bart Bussmann, Patrick Leask, Javier Ferrando, Oscar Obeso, and Stepan Shabalin for the valuable input during MATS. We thank the entire MATS and Lighthaven staff for creating the environment that made thisresearch possible. Lastly, we are thankful for computational resources provided by David Bau's lab.

\bibliographystyle{abbrvnat}
\bibliography{neurips_2024}

\begin{thebibliography}{31}
\providecommand{\natexlab}[1]{#1}
\providecommand{\url}[1]{\texttt{#1}}
\expandafter\ifx\csname urlstyle\endcsname\relax
  \providecommand{\doi}[1]{doi: #1}\else
  \providecommand{\doi}{doi: \begingroup \urlstyle{rm}\Url}\fi

\bibitem[Belrose et~al.(2023)Belrose, Schneider-Joseph, Ravfogel, Cotterell, Raff, and Biderman]{belrose2023leace}
N.~Belrose, D.~Schneider-Joseph, S.~Ravfogel, R.~Cotterell, E.~Raff, and S.~Biderman.
\newblock {LEACE}: Perfect linear concept erasure in closed form.
\newblock In \emph{Thirty-seventh Conference on Neural Information Processing Systems}, 2023.
\newblock URL \url{https://openreview.net/forum?id=awIpKpwTwF}.

\bibitem[Biderman et~al.(2023)Biderman, Schoelkopf, Anthony, Bradley, O'Brien, Hallahan, Khan, Purohit, Prashanth, Raff, Skowron, Sutawika, and van~der Wal]{biderman2023pythiasuiteanalyzinglarge}
S.~Biderman, H.~Schoelkopf, Q.~Anthony, H.~Bradley, K.~O'Brien, E.~Hallahan, M.~A. Khan, S.~Purohit, U.~S. Prashanth, E.~Raff, A.~Skowron, L.~Sutawika, and O.~van~der Wal.
\newblock Pythia: A suite for analyzing large language models across training and scaling, 2023.
\newblock URL \url{https://arxiv.org/abs/2304.01373}.

\bibitem[Bolukbasi et~al.(2016)Bolukbasi, Chang, Zou, Saligrama, and Kalai]{bolukbasi2016man}
T.~Bolukbasi, K.-W. Chang, J.~Y. Zou, V.~Saligrama, and A.~T. Kalai.
\newblock Man is to computer programmer as woman is to homemaker? debiasing word embeddings.
\newblock In \emph{Advances in Neural Information Processing Systems}, volume~29, pages 4349--4357, 2016.

\bibitem[Braun et~al.(2024)Braun, Taylor, Goldowsky-Dill, and Sharkey]{braun2024identifyingfunctionallyimportantfeatures}
D.~Braun, J.~Taylor, N.~Goldowsky-Dill, and L.~Sharkey.
\newblock Identifying functionally important features with end-to-end sparse dictionary learning, 2024.
\newblock URL \url{https://arxiv.org/abs/2405.12241}.

\bibitem[Bricken et~al.(2023)Bricken, Templeton, Batson, Chen, Jermyn, Conerly, Turner, Anil, Denison, Askell, Lasenby, Wu, Kravec, Schiefer, Maxwell, Joseph, Hatfield-Dodds, Tamkin, Nguyen, McLean, Burke, Hume, Carter, Henighan, and Olah]{bricken2023monosemanticity}
T.~Bricken, A.~Templeton, J.~Batson, B.~Chen, A.~Jermyn, T.~Conerly, N.~Turner, C.~Anil, C.~Denison, A.~Askell, R.~Lasenby, Y.~Wu, S.~Kravec, N.~Schiefer, T.~Maxwell, N.~Joseph, Z.~Hatfield-Dodds, A.~Tamkin, K.~Nguyen, B.~McLean, J.~E. Burke, T.~Hume, S.~Carter, T.~Henighan, and C.~Olah.
\newblock Towards monosemanticity: Decomposing language models with dictionary learning.
\newblock \emph{Transformer Circuits Thread}, 2023.
\newblock https://transformer-circuits.pub/2023/monosemantic-features/index.html.

\bibitem[Bussmann et~al.(2024)Bussmann, Leask, and Nanda]{bussmann2024batchtopk}
B.~Bussmann, P.~Leask, and N.~Nanda.
\newblock Batchtopk: A simple improvement for topk-saes.
\newblock \emph{AI Alignment Forum}, Jul 2024.
\newblock URL \url{https://www.lesswrong.com/posts/Nkx6yWZNbAsfvic98/batchtopk-a-simple-improvement-for-topk-saes}.

\bibitem[Cunningham et~al.(2023)Cunningham, Ewart, Riggs, Huben, and Sharkey]{cunningham2023sparseautoencodershighlyinterpretable}
H.~Cunningham, A.~Ewart, L.~Riggs, R.~Huben, and L.~Sharkey.
\newblock Sparse autoencoders find highly interpretable features in language models, 2023.
\newblock URL \url{https://arxiv.org/abs/2309.08600}.

\bibitem[De-Arteaga et~al.(2019)De-Arteaga, Romanov, Wallach, Chayes, Borgs, Chouldechova, Geyik, Kenthapadi, and Kalai]{De_Arteaga_2019}
M.~De-Arteaga, A.~Romanov, H.~Wallach, J.~Chayes, C.~Borgs, A.~Chouldechova, S.~Geyik, K.~Kenthapadi, and A.~T. Kalai.
\newblock Bias in bios: A case study of semantic representation bias in a high-stakes setting.
\newblock In \emph{Proceedings of the Conference on Fairness, Accountability, and Transparency}, FAT* ’19. ACM, Jan. 2019.
\newblock \doi{10.1145/3287560.3287572}.
\newblock URL \url{http://dx.doi.org/10.1145/3287560.3287572}.

\bibitem[Elhage et~al.(2022)Elhage, Hume, Olsson, Schiefer, Henighan, Kravec, Hatfield-Dodds, Lasenby, Drain, Chen, Grosse, McCandlish, Kaplan, Amodei, Wattenberg, and Olah]{elhage2022superposition}
N.~Elhage, T.~Hume, C.~Olsson, N.~Schiefer, T.~Henighan, S.~Kravec, Z.~Hatfield-Dodds, R.~Lasenby, D.~Drain, C.~Chen, R.~Grosse, S.~McCandlish, J.~Kaplan, D.~Amodei, M.~Wattenberg, and C.~Olah.
\newblock Toy models of superposition.
\newblock \emph{Transformer Circuits Thread}, 2022.
\newblock https://transformer-circuits.pub/2022/toy\_model/index.html.

\bibitem[Ferrando et~al.(2024)Ferrando, Sarti, Bisazza, and Costa-jussà]{ferrando2024primerinnerworkingstransformerbased}
J.~Ferrando, G.~Sarti, A.~Bisazza, and M.~R. Costa-jussà.
\newblock A primer on the inner workings of transformer-based language models, 2024.
\newblock URL \url{https://arxiv.org/abs/2405.00208}.

\bibitem[Gao et~al.(2020)Gao, Biderman, Black, Golding, Hoppe, Foster, Phang, He, Thite, Nabeshima, Presser, and Leahy]{gao2020pile800gbdatasetdiverse}
L.~Gao, S.~Biderman, S.~Black, L.~Golding, T.~Hoppe, C.~Foster, J.~Phang, H.~He, A.~Thite, N.~Nabeshima, S.~Presser, and C.~Leahy.
\newblock The pile: An 800gb dataset of diverse text for language modeling, 2020.
\newblock URL \url{https://arxiv.org/abs/2101.00027}.

\bibitem[Gao et~al.(2024)Gao, la~Tour, Tillman, Goh, Troll, Radford, Sutskever, Leike, and Wu]{gao2024scalingevaluatingsparseautoencoders}
L.~Gao, T.~D. la~Tour, H.~Tillman, G.~Goh, R.~Troll, A.~Radford, I.~Sutskever, J.~Leike, and J.~Wu.
\newblock Scaling and evaluating sparse autoencoders, 2024.
\newblock URL \url{https://arxiv.org/abs/2406.04093}.

\bibitem[Hou et~al.(2024)Hou, Li, He, Yan, Chen, and McAuley]{hou2024bridging}
Y.~Hou, J.~Li, Z.~He, A.~Yan, X.~Chen, and J.~McAuley.
\newblock Bridging language and items for retrieval and recommendation.
\newblock \emph{arXiv preprint arXiv:2403.03952}, 2024.

\bibitem[Iskander et~al.(2023)Iskander, Radinsky, and Belinkov]{iskander-etal-2023-shielded}
S.~Iskander, K.~Radinsky, and Y.~Belinkov.
\newblock Shielded representations: Protecting sensitive attributes through iterative gradient-based projection.
\newblock In A.~Rogers, J.~Boyd-Graber, and N.~Okazaki, editors, \emph{Findings of the Association for Computational Linguistics: ACL 2023}, pages 5961--5977, Toronto, Canada, July 2023. Association for Computational Linguistics.
\newblock \doi{10.18653/v1/2023.findings-acl.369}.
\newblock URL \url{https://aclanthology.org/2023.findings-acl.369}.

\bibitem[Jermyn et~al.(2024)Jermyn, Templeton, Batson, and Bricken]{jermyn2024tanh}
A.~Jermyn, A.~Templeton, J.~Batson, and T.~Bricken.
\newblock Tanh penalty in dictionary learning.
\newblock \url{https://transformer-circuits.pub/2024/feb-update/index.html#dict-learning-tanh}, Feb 2024.
\newblock In: Circuits Updates - February 2024.

\bibitem[Juang et~al.(2024)Juang, Paulo, Drori, and Belrose]{juang2024opensourceautointerp}
C.~Juang, G.~Paulo, J.~Drori, and N.~Belrose.
\newblock Open source automated interpretability for sparse autoencoder features, jul 2024.
\newblock URL \url{https://blog.eleuther.ai/autointerp/}.
\newblock Building and evaluating an open-source pipeline for auto-interpretability.

\bibitem[Karvonen et~al.(2024)Karvonen, Wright, Rager, Angell, Brinkmann, Smith, Verdun, Bau, and Marks]{karvonen2024measuringprogressdictionarylearning}
A.~Karvonen, B.~Wright, C.~Rager, R.~Angell, J.~Brinkmann, L.~Smith, C.~M. Verdun, D.~Bau, and S.~Marks.
\newblock Measuring progress in dictionary learning for language model interpretability with board game models, 2024.
\newblock URL \url{https://arxiv.org/abs/2408.00113}.

\bibitem[Lieberum et~al.(2024)Lieberum, Rajamanoharan, Conmy, Smith, Sonnerat, Varma, Kramár, Dragan, Shah, and Nanda]{lieberum2024gemmascopeopensparse}
T.~Lieberum, S.~Rajamanoharan, A.~Conmy, L.~Smith, N.~Sonnerat, V.~Varma, J.~Kramár, A.~Dragan, R.~Shah, and N.~Nanda.
\newblock Gemma scope: Open sparse autoencoders everywhere all at once on gemma 2, 2024.
\newblock URL \url{https://arxiv.org/abs/2408.05147}.

\bibitem[Makelov et~al.(2024)Makelov, Lange, and Nanda]{makelov2024subspace}
A.~Makelov, G.~Lange, and N.~Nanda.
\newblock Is this the subspace you are looking for? an interpretability illusion for subspace activation patching.
\newblock In \emph{The Twelfth International Conference on Learning Representations}, 2024.
\newblock URL \url{https://openreview.net/forum?id=Ebt7JgMHv1}.

\bibitem[Marks et~al.(2024)Marks, Rager, Michaud, Belinkov, Bau, and Mueller]{marks2024sparsefeaturecircuitsdiscovering}
S.~Marks, C.~Rager, E.~J. Michaud, Y.~Belinkov, D.~Bau, and A.~Mueller.
\newblock Sparse feature circuits: Discovering and editing interpretable causal graphs in language models, 2024.
\newblock URL \url{https://arxiv.org/abs/2403.19647}.

\bibitem[Nanda(2023)]{nanda2023attribution}
N.~Nanda.
\newblock Attribution patching: Activation patching at industrial scale, 2023.
\newblock URL \url{https://www.neelnanda.io/mechanistic-interpretability/attribution-patching}.

\bibitem[nostalgebraist(2020)]{nostalgebraist2020interpreting}
nostalgebraist.
\newblock Interpreting gpt: the logit lens, 2020.
\newblock URL \url{https://www.lesswrong.com/posts/AcKRB8wDpdaN6v6ru/interpreting-gpt-the-logit-lens}.

\bibitem[Rajamanoharan et~al.(2024{\natexlab{a}})Rajamanoharan, Conmy, Smith, Lieberum, Varma, Kramár, Shah, and Nanda]{rajamanoharan2024improvingdictionarylearninggated}
S.~Rajamanoharan, A.~Conmy, L.~Smith, T.~Lieberum, V.~Varma, J.~Kramár, R.~Shah, and N.~Nanda.
\newblock Improving dictionary learning with gated sparse autoencoders, 2024{\natexlab{a}}.
\newblock URL \url{https://arxiv.org/abs/2404.16014}.

\bibitem[Rajamanoharan et~al.(2024{\natexlab{b}})Rajamanoharan, Lieberum, Sonnerat, Conmy, Varma, Kramár, and Nanda]{rajamanoharan2024jumpingaheadimprovingreconstruction}
S.~Rajamanoharan, T.~Lieberum, N.~Sonnerat, A.~Conmy, V.~Varma, J.~Kramár, and N.~Nanda.
\newblock Jumping ahead: Improving reconstruction fidelity with jumprelu sparse autoencoders, 2024{\natexlab{b}}.
\newblock URL \url{https://arxiv.org/abs/2407.14435}.

\bibitem[Ravfogel et~al.(2020)Ravfogel, Elazar, Gonen, Twiton, and Goldberg]{ravfogel-etal-2020-null}
S.~Ravfogel, Y.~Elazar, H.~Gonen, M.~Twiton, and Y.~Goldberg.
\newblock Null it out: Guarding protected attributes by iterative nullspace projection.
\newblock In D.~Jurafsky, J.~Chai, N.~Schluter, and J.~Tetreault, editors, \emph{Proceedings of the 58th Annual Meeting of the Association for Computational Linguistics}, pages 7237--7256, Online, July 2020. Association for Computational Linguistics.
\newblock \doi{10.18653/v1/2020.acl-main.647}.
\newblock URL \url{https://aclanthology.org/2020.acl-main.647}.

\bibitem[Ravfogel et~al.(2022{\natexlab{a}})Ravfogel, Twiton, Goldberg, and Cotterell]{pmlr-v162-ravfogel22a}
S.~Ravfogel, M.~Twiton, Y.~Goldberg, and R.~D. Cotterell.
\newblock Linear adversarial concept erasure.
\newblock In K.~Chaudhuri, S.~Jegelka, L.~Song, C.~Szepesvari, G.~Niu, and S.~Sabato, editors, \emph{Proceedings of the 39th International Conference on Machine Learning}, volume 162 of \emph{Proceedings of Machine Learning Research}, pages 18400--18421. PMLR, 17--23 Jul 2022{\natexlab{a}}.
\newblock URL \url{https://proceedings.mlr.press/v162/ravfogel22a.html}.

\bibitem[Ravfogel et~al.(2022{\natexlab{b}})Ravfogel, Vargas, Goldberg, and Cotterell]{ravfogel-etal-2022-adversarial}
S.~Ravfogel, F.~Vargas, Y.~Goldberg, and R.~Cotterell.
\newblock Adversarial concept erasure in kernel space.
\newblock In Y.~Goldberg, Z.~Kozareva, and Y.~Zhang, editors, \emph{Proceedings of the 2022 Conference on Empirical Methods in Natural Language Processing}, pages 6034--6055, Abu Dhabi, United Arab Emirates, Dec. 2022{\natexlab{b}}. Association for Computational Linguistics.
\newblock \doi{10.18653/v1/2022.emnlp-main.405}.
\newblock URL \url{https://aclanthology.org/2022.emnlp-main.405}.

\bibitem[Sharkey et~al.(2022)Sharkey, Braun, and beren]{sharkey2022interim}
L.~Sharkey, D.~Braun, and beren.
\newblock [interim research report] taking features out of superposition with sparse autoencoders, 12 2022.
\newblock URL \url{https://www.lesswrong.com/posts/z6QQJbtpkEAX3Aojj/interim-research-report-taking-features-out-of-superposition}.
\newblock AI Alignment Forum.

\bibitem[Team(2024)]{gemmateam2024gemma2improvingopen}
G.~Team.
\newblock Gemma 2: Improving open language models at a practical size, 2024.
\newblock URL \url{https://arxiv.org/abs/2408.00118}.

\bibitem[Templeton et~al.(2024)Templeton, Conerly, Marcus, Lindsey, Bricken, Chen, Pearce, Citro, Ameisen, Jones, Cunningham, Turner, McDougall, MacDiarmid, Freeman, Sumers, Rees, Batson, Jermyn, Carter, Olah, and Henighan]{templeton2024scaling}
A.~Templeton, T.~Conerly, J.~Marcus, J.~Lindsey, T.~Bricken, B.~Chen, A.~Pearce, C.~Citro, E.~Ameisen, A.~Jones, H.~Cunningham, N.~L. Turner, C.~McDougall, M.~MacDiarmid, C.~D. Freeman, T.~R. Sumers, E.~Rees, J.~Batson, A.~Jermyn, S.~Carter, C.~Olah, and T.~Henighan.
\newblock Scaling monosemanticity: Extracting interpretable features from claude 3 sonnet.
\newblock \emph{Transformer Circuits Thread}, 2024.
\newblock URL \url{https://transformer-circuits.pub/2024/scaling-monosemanticity/index.html}.

\bibitem[Wang et~al.(2020)Wang, Lin, Rajani, McCann, Ordonez, and Xiong]{wang-etal-2020-double}
T.~Wang, X.~V. Lin, N.~F. Rajani, B.~McCann, V.~Ordonez, and C.~Xiong.
\newblock Double-hard debias: Tailoring word embeddings for gender bias mitigation.
\newblock In D.~Jurafsky, J.~Chai, N.~Schluter, and J.~Tetreault, editors, \emph{Proceedings of the 58th Annual Meeting of the Association for Computational Linguistics}, pages 5443--5453, Online, July 2020. Association for Computational Linguistics.
\newblock \doi{10.18653/v1/2020.acl-main.484}.
\newblock URL \url{https://aclanthology.org/2020.acl-main.484}.

\end{thebibliography}

\clearpage
\appendix
\section{Appendix}
\subsection{Probe Training}
\label{apx:probe}

When training probes on biased datasets (e.g., male\_professor, female\_nurse), the probe can potentially leverage two distinct signals: profession and gender. The probe must trade off between these signals. Therefore, evaluating the biased probe on balanced gender and profession datasets will result in a mean accuracy of at most 75\%.

For effective spurious correlation removal and differentiation between SAEs, the probe must capture both signals to some degree. If the probe's accuracy is heavily skewed (e.g., 50\% on class 1 and 95\% on class 2), ablating SAE latents corresponding to class 2 cannot significantly improve class 1 accuracy, limiting the effectiveness of the evaluation.

We find that the direction of the spurious correlation varies based on the underlying LLM selected. The strength of this spurious correlation is influenced by probe training hyperparameters. For instance, using a large probe batch size on Gemma-2-2B resulted in a weak spurious correlation (< 55\% accuracy) for the gender category. Decreasing the batch size led to the probe learning a more balanced mix of both signals, enabling significant improvements in probe accuracy on a desired concept and better differentiation between SAEs. In addition, we find that a high learning rate introduces variance in the direction and strength of the spurious correlation.

In our experiments with Pythia-70M, a smaller model, we found that the probe mostly relied on the gender signal. This scenario provides an intuitive demonstration of removing unwanted gender bias, where we ablate gender features to improve profession classification. However, with larger models like Gemma-2-2B, we observed that our probes primarily picked up on profession signals. In such cases, to significantly change probe accuracy and differentiate between SAEs, we need to ablate the profession signal to improve gender classification. 

To decrease the dependence of the SHIFT on our selection of individual dataset classes, we average scores over multiple class pairs. We consider the pairs ("professor', 'nurse") and ("architect', 'journalist") in the Bias and Bios dataset, and the pairs ("Books", "CDs and Vinyl"), ("Software", "Electronics"), ("Pet Supplies", "Office Products"), ("Industrial and Scientific", "Toys and Games"). These classes were selected based on the objective that a classifier $C_\text{b}$ develops an accuracy of at least 0.6 for both categories in the pair (gender / profession or sentiment / product category).

\begin{table}[ht]
\centering
\begin{minipage}[t]{0.4\textwidth}
    \centering
    \resizebox{\textwidth}{!}{%
    \begin{tabular}{>{\raggedright\arraybackslash}p{3cm} c}
    \toprule
    \textbf{Parameter} & \textbf{Value} \\
    \midrule
    LLM Context Length & 128 \\
    Input Datapoints & 4000 \\
    Epochs & 5 \\
    Optimizer & \texttt{Adam} \\
    Adam betas & (0.9, 0.999) \\
    Batch size & 16 \\
    Learning rate & 1e-3 \\
    \bottomrule
    \end{tabular}
    }
    \vspace{0.2cm}
    \caption{Training parameters of our linear probes.}
    \label{tab:probe-hyperparameters}
\end{minipage}%
\hfill
\begin{minipage}[t]{0.48\textwidth}
    \centering
    \resizebox{\textwidth}{!}{%
    \begin{tabular}{>{\raggedright\arraybackslash}p{2.2cm} *{4}{c}}
    \toprule
    & \multicolumn{2}{c}{Clean} & \multicolumn{2}{c}{Mod.} \\
    \cmidrule(lr){2-3} \cmidrule(lr){4-5}
    \textbf{Model} & \textbf{Gen.} & \textbf{Prof.} & \textbf{Gen.} & \textbf{Prof.} \\
    \midrule
    \textbf{Pythia-70M} \\
    \midrule
    Prof. / Nurse & 0.77 & 0.72 & 0.52 & 0.91 \\
    Arch. / Journ. & 0.86 & 0.63 & 0.53 & 0.88 \\
    \midrule
    \textbf{Gemma-2-2B} \\
    \midrule
    Prof. / Nurse & 0.64 & 0.86 & 0.83 & 0.65 \\
    Arch. / Journ. & 0.69 & 0.80 & 0.94 & 0.55 \\
    \bottomrule
    \end{tabular}
    }
    \vspace{0.2cm}
    \caption{Original and modified accuracies of biased probes evaluated on balanced gender and profession datasets. Modified accuracies represent the best accuracies obtained using the SHIFT method on any SAE.}
    \label{tab:scr_probe_accs}
\end{minipage}
\end{table}

\clearpage
\subsection{Sparse Autoencoder Training}
\label{apx:sae}

As a testbed for our evaluation pipeline, we train and open-source a suite of SAEs on the models pythia-70m-deduped \cite{biderman2023pythiasuiteanalyzinglarge} and Gemma-2-2B \cite{gemmateam2024gemma2improvingopen}. We train Vanilla and TopK architectures on 200M tokens from The Pile \cite{gao2020pile800gbdatasetdiverse} with expansion factors 8x (Pythia, Gemma) and 32x (Pythia).
\footnote{The expansion factor denotes the ratio of SAE latents to input dimension.} Additionally, we evaluate JumpReLU SAEs from the Gemma-Scope suite \cite{lieberum2024gemmascopeopensparse}.


\begin{table}[htbp]
    \centering
    \caption{Training parameters of our sparse autoencoders.}
    \vspace{0.2cm}
    \label{tab:sae-hyperparameters}
    \begin{tabular}{l|c}
        \hline
        \textbf{Parameter} & \textbf{Value} \\ \hline
        LLM Context Length & 128 \\
        Number of tokens & 200M \\
        Optimizer & \texttt{Adam} \\ 
        Adam betas & (0.9, 0.999) \\ 
        Linear warmup steps & 1,000 \\ 
        Batch size & 4,096 \\ 
        Learning rate & 3e-4 \\ 
        Expansion factor & \{8, 32\} \\ 
        \hline
    \end{tabular}
\end{table}

\clearpage
\subsection{Gemma Results without Auto-interp}

\begin{figure}[htb!]
\centering
\begin{tabular}{cc}
\includegraphics[width=0.49\textwidth]{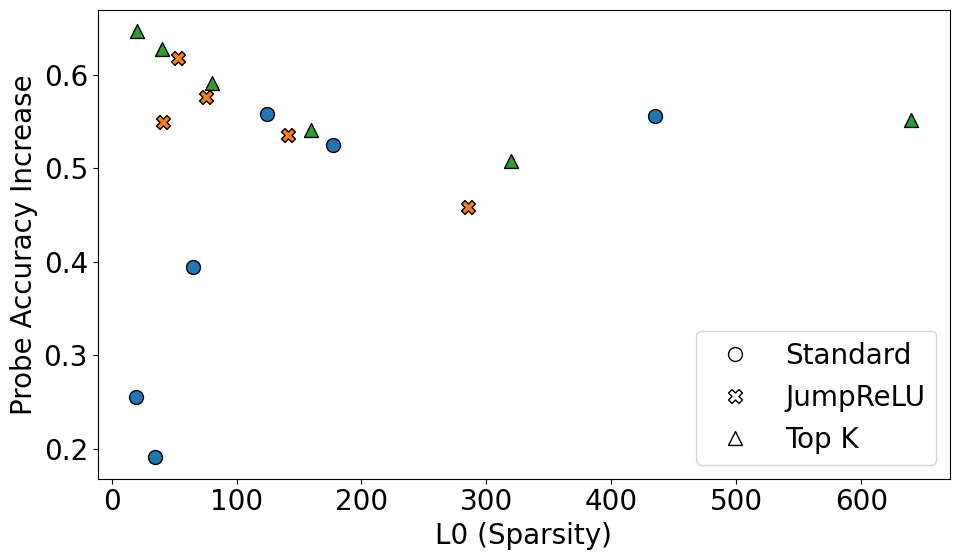} &
\includegraphics[width=0.49\textwidth]{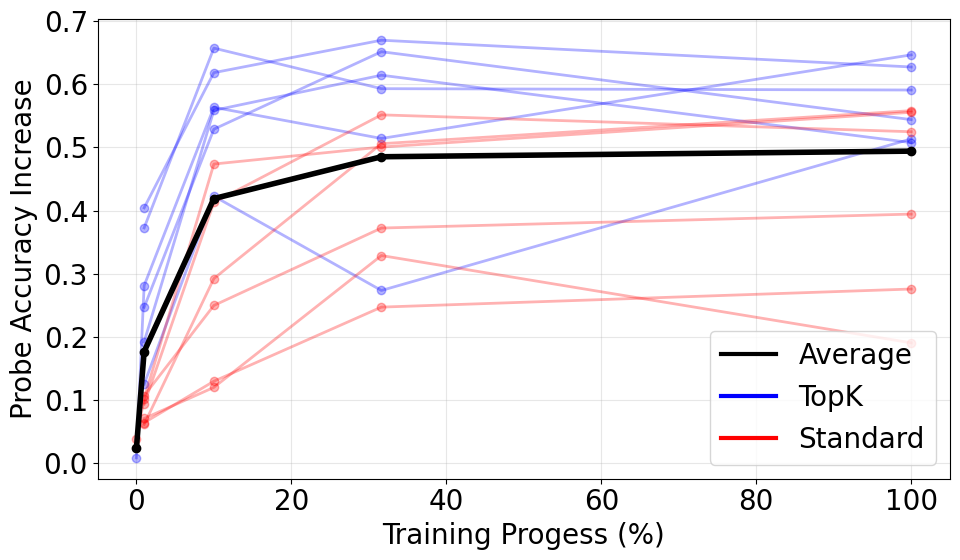}
\end{tabular}
\caption{ SCR scores without auto-interp as a function of training progress, measured for checkpoints at 0\%, 1\%, 10\%, 31\%, and 100\% of SAE training over 6 TopK and 6 Standard SAEs (right). Each datapoint (left) and line (right) corresponds to a single SAE, architectures are differentiated by color.
}\label{fig:gemma_tpp_no_autointerp_plots}
\end{figure}

For TPP without an LLM judge, we find that a randomly initialized Standard SAE performs relatively well on our metric in Figure \ref{fig:gemma_tpp_no_autointerp_plots}. We believe this is because the randomly initialized SAE has dense activations with an L0 of 9000, significantly larger than Gemma-2-2B's d\_model of 2304. Once training begins and the Standard SAE becomes more sparse, the results with and without the LLM judge begin to correlate.

\begin{figure}[htb!]
\centering
\begin{tabular}{cc}
\includegraphics[width=0.49\textwidth]{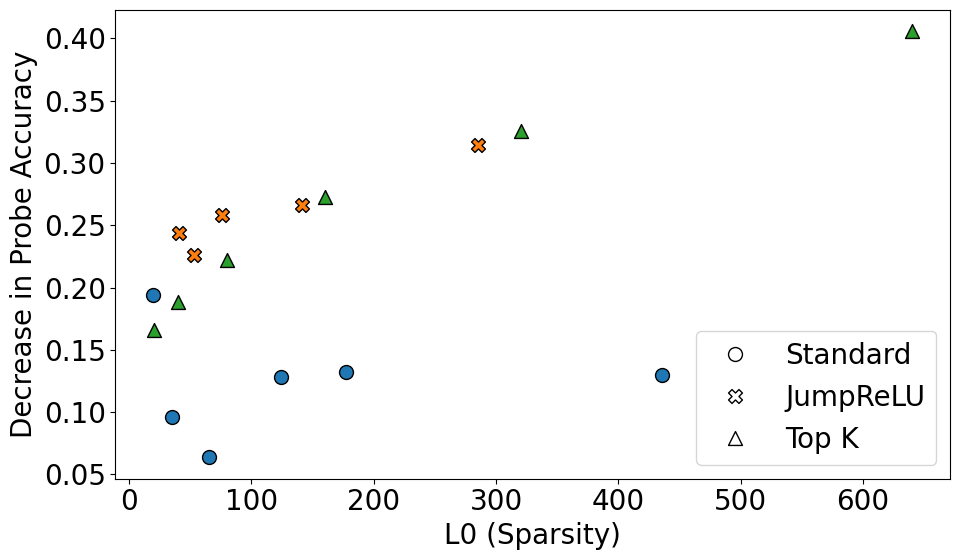} &
\includegraphics[width=0.49\textwidth]{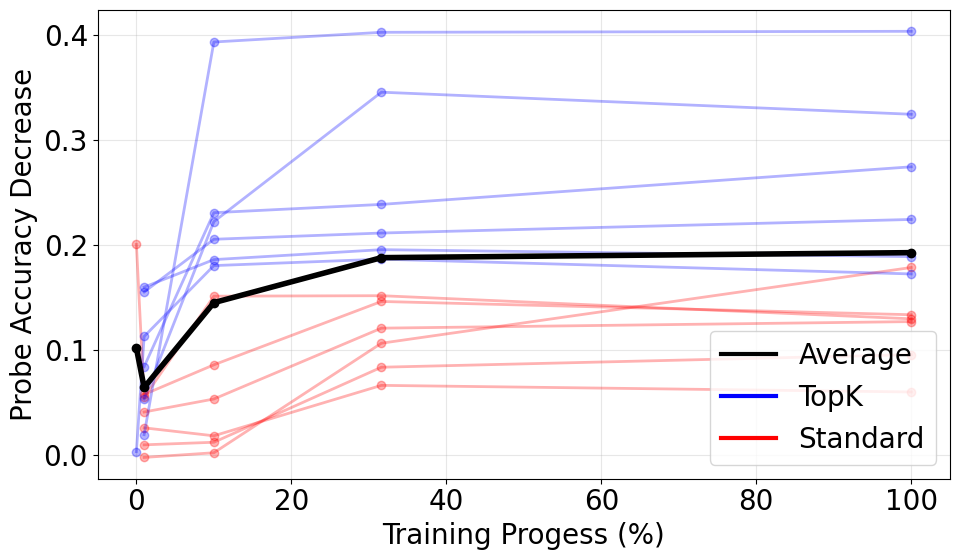}
\end{tabular}
\caption{ Targeted Probe Perturbation (TPP) scores without auto-interp over sparsity for SAEs of Standard, JumpReLU, and TopK architectures (left). TPP scores as a function of training progress, measured for checkpoints at 0\%, 1\%, 10\%, 31\%, and 100\% of SAE training over 6 TopK and 6 Standard SAEs (right). Each datapoint (left) and line (right) corresponds to a single SAE, architectures are differentiated by color.
}\label{fig:gemma_scr_no_autointerp_plots}
\end{figure}

\clearpage
\subsection{Pythia-70M Results}
\label{apx:pythia}

\begin{figure}[htb!]
\centering
\begin{subfigure}[b]{0.49\textwidth}
    \includegraphics[width=\textwidth]{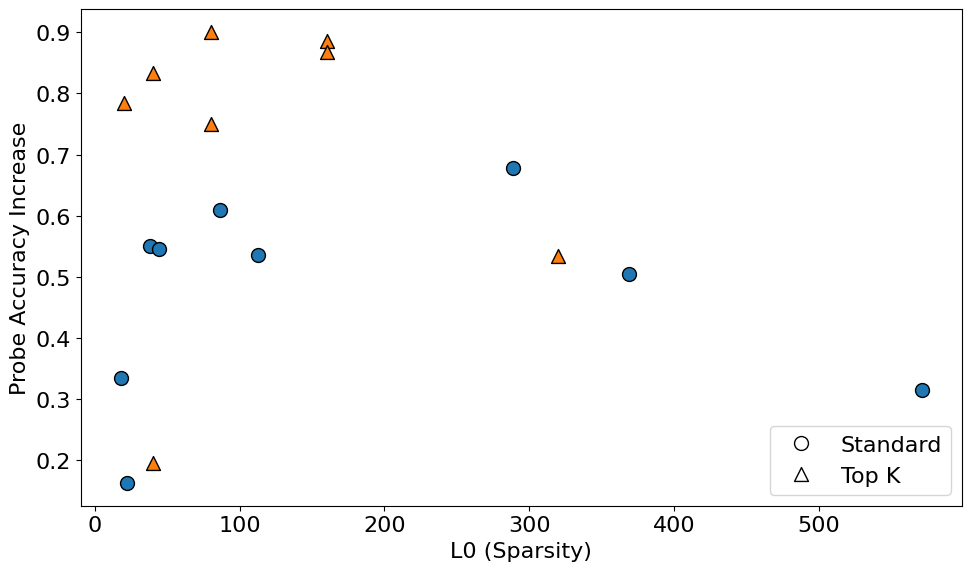}
    \label{fig:sub1pythia_plots}
    \caption{SCR, No Auto interp}
\end{subfigure}
\hfill 
\begin{subfigure}[b]{0.49\textwidth}
    \includegraphics[width=\textwidth]{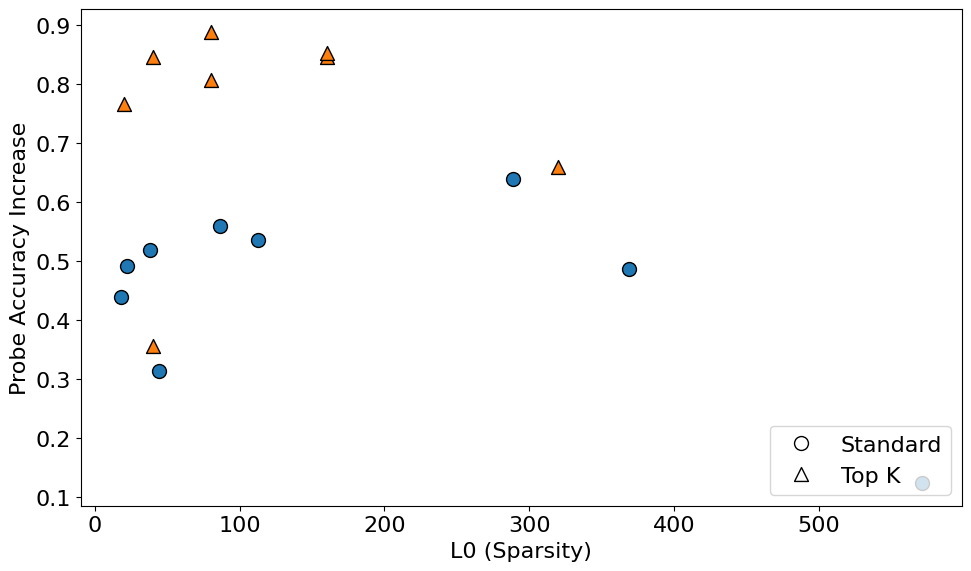}
    \label{fig:sub2pythia_plots}
    \caption{SCR, Auto interp}
\end{subfigure}

\begin{subfigure}[b]{0.49\textwidth}
    \includegraphics[width=\textwidth]{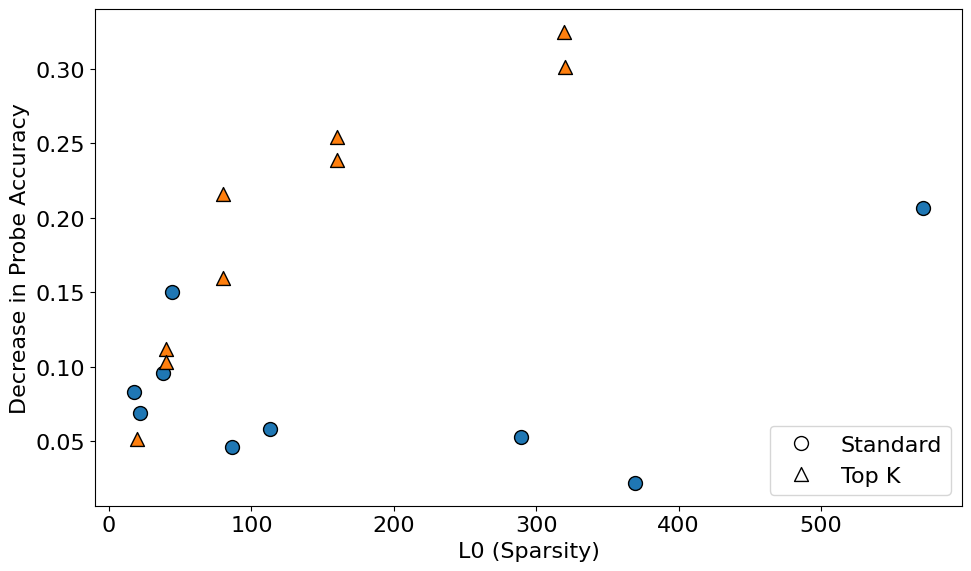}
    \label{fig:sub3pythia_plots}
    \caption{TPP, No Auto interp}
\end{subfigure}
\hfill
\begin{subfigure}[b]{0.49\textwidth}
    \includegraphics[width=\textwidth]{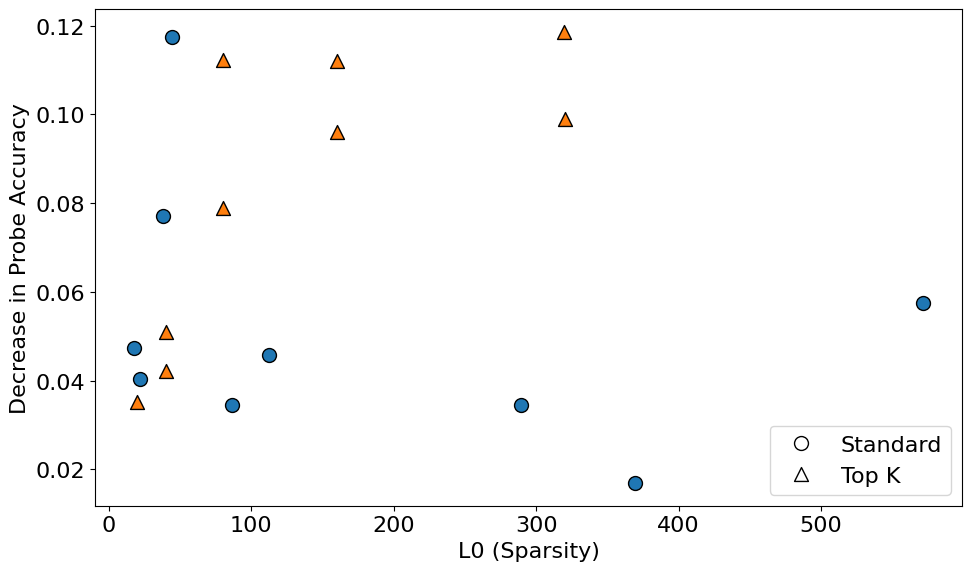}
    \label{fig:sub4pythia_plots}
    \caption{TPP, Auto interp}
\end{subfigure}

\caption{Results for Pythia-70M. The left column contains a scatterplot of loss recovered vs L0, with color corresponding to coverage score, and each point representing different hyperparameters. We differentiate between SAE training methods with shapes.
}\label{fig:pythia_plots}
\end{figure}

Complementary to our results in Section \ref{sec:results} we provide an evaluation sweep for SHIFT and TPP on the Pythia-70M model in Figure \ref{fig:pythia_plots}. Additionally we compare SAE architectures in the number of SAEs required to achieve a high SHIFT score in Table \ref{tab:shift_comp}.

\begin{table}[htbp]
    \centering
    \caption{SAE intervention locations and their Ground Truth Accuracy.}
    \vspace{0.2cm}
    \label{tab:sae-intervention-locations}
    \begin{tabular}{l|c}
        \hline
        \textbf{SAE intervention locations} & \textbf{Ground Truth Acc} \\ \hline
        None & 59\% \\
        Standard SAE, Embedding and Layers 0-4 & 88\% \\
        Standard SAE, Embedding only & 86\% \\
        Standard SAE, Layers 0-4 & 84\% \\
        Standard SAE, Layers 3-4 & 81\% \\
        Standard SAE, Layers 3-4, Resid only & 79\% \\
        TopK SAE, Layer 4, Resid only & \textbf{90\%} \\
        \hline
    \end{tabular}
\caption{ TopK SAEs significantly improve performance of the SHIFT method in the setting used in Marks et al. \cite{marks2024sparsefeaturecircuitsdiscovering}. A biased probe is trained on the class pair of ("professor", "nurse"). By ablating gender-related features, we improve the probe's accuracy at profession classification. In Marks et al., 16 Standard SAEs were used on MLP output, attention output, and resid\_post in layers 0-4, in addition to the embedding output. We exceed their performance using only a single TopK SAE.}
\label{tab:shift_comp}
\end{table}

\clearpage
\subsection{Amazon SCR Results}
\label{apx:amazon_scr}

In the Bias in Bios dataset, we evaluate the concepts of profession / gender. In the Amazon reviews dataset, the corresponding concepts are product category / review sentiment.

We did not evaluate SCR or TPP on Pythia-70M with the Amazon dataset, as we found that probes trained on Pythia-70M obtained poor scores for identifying Product Category.

We evaluate the following product category pairs:     ("Books","CDs\_and\_Vinyl"), ("Software", "Electronics"), ("Pet\_Supplies", "Office\_Products"). 

\begin{figure}[htb!]
\centering
\begin{subfigure}[b]{0.49\textwidth}
    \includegraphics[width=\textwidth]{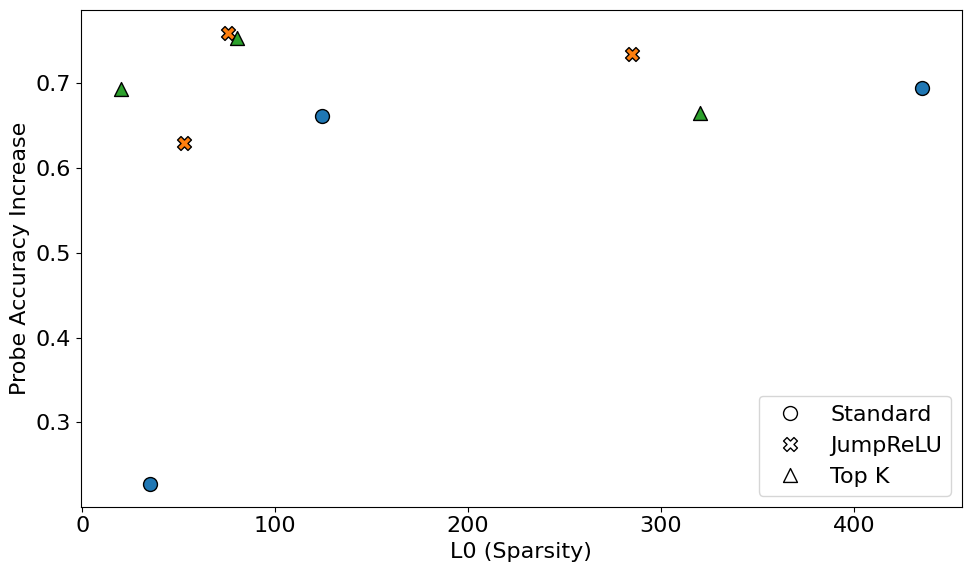}
    \label{fig:sub1amazon_scr_plots}
    \caption{SCR, No Auto interp}
\end{subfigure}
\hfill 
\begin{subfigure}[b]{0.49\textwidth}
    \includegraphics[width=\textwidth]{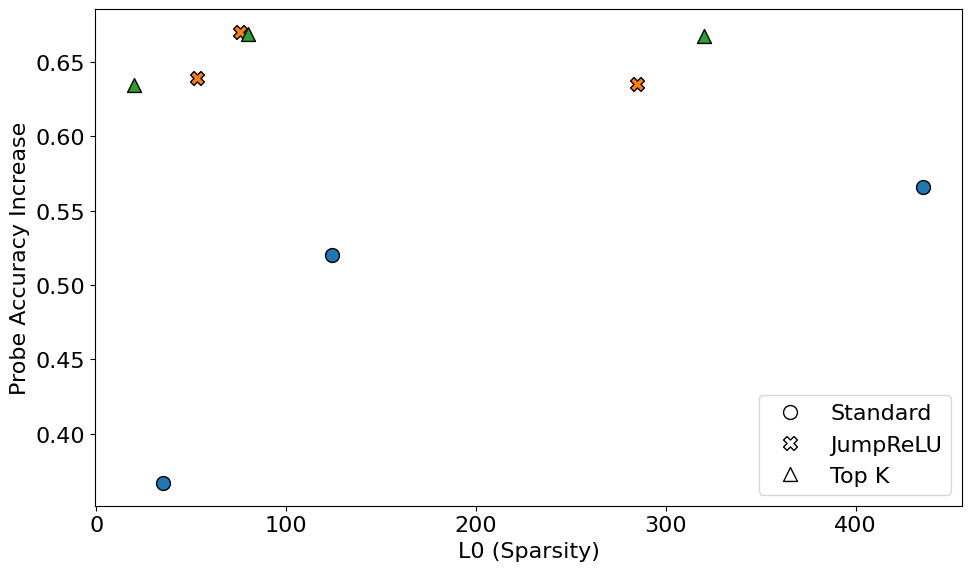}
    \label{fig:sub2amazon_scr_plots}
    \caption{SCR, Auto interp}
\end{subfigure}

\caption{Results for Amazon SCR, Gemma-2-2B. The left column contains a scatterplot of loss recovered vs L0, with color corresponding to coverage score, and each point representing different hyperparameters. We differentiate between SAE training methods with shapes.
}\label{fig:amazon_scr_plots}
\end{figure}

\clearpage
\subsection{Correlation of SHIFT scores with interpretability.}
\label{apx:judge_corr}

\begin{figure}[htb!]
    \centering
    \includegraphics[width=0.49\linewidth]{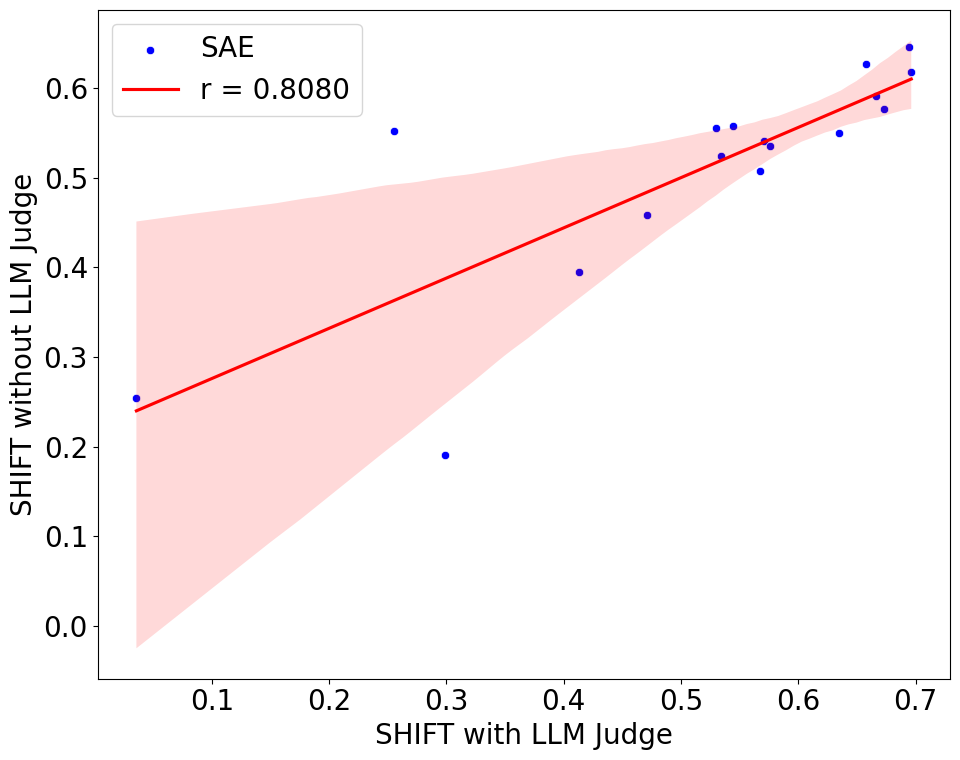}
    \includegraphics[width=0.49\linewidth]{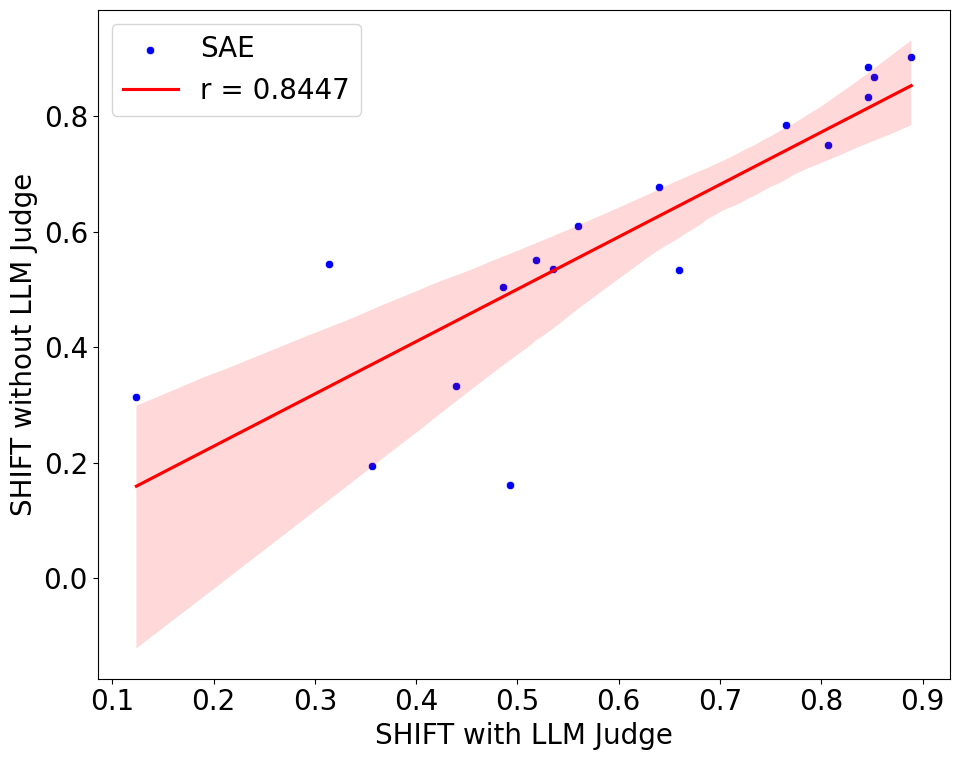}
    \caption{Correlation of SHIFT scores on Gemma-2-2B (left) and Pythia-70M before and after filtering SAE latents with the LLM judge described in Section \ref{sec:method:latent_selection}. The red area denotes the 95\% confidence interval.}
    \label{fig:scr_corr}
\end{figure}

\begin{figure}[htb!]
    \centering
    \includegraphics[width=0.49\linewidth]{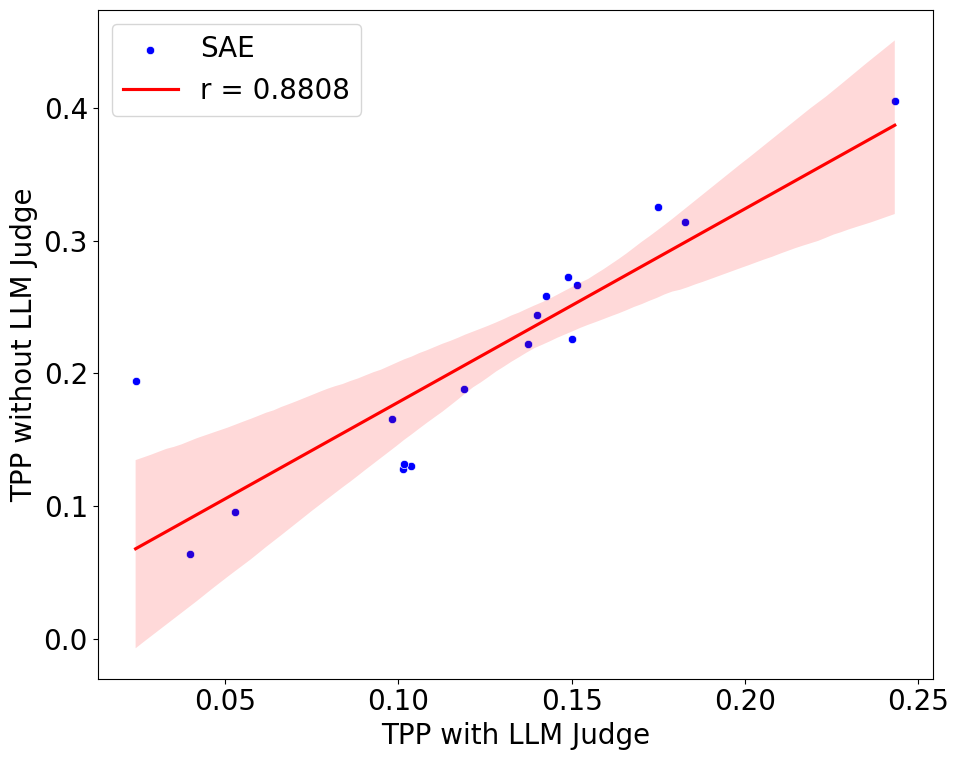}
    \includegraphics[width=0.49\linewidth]{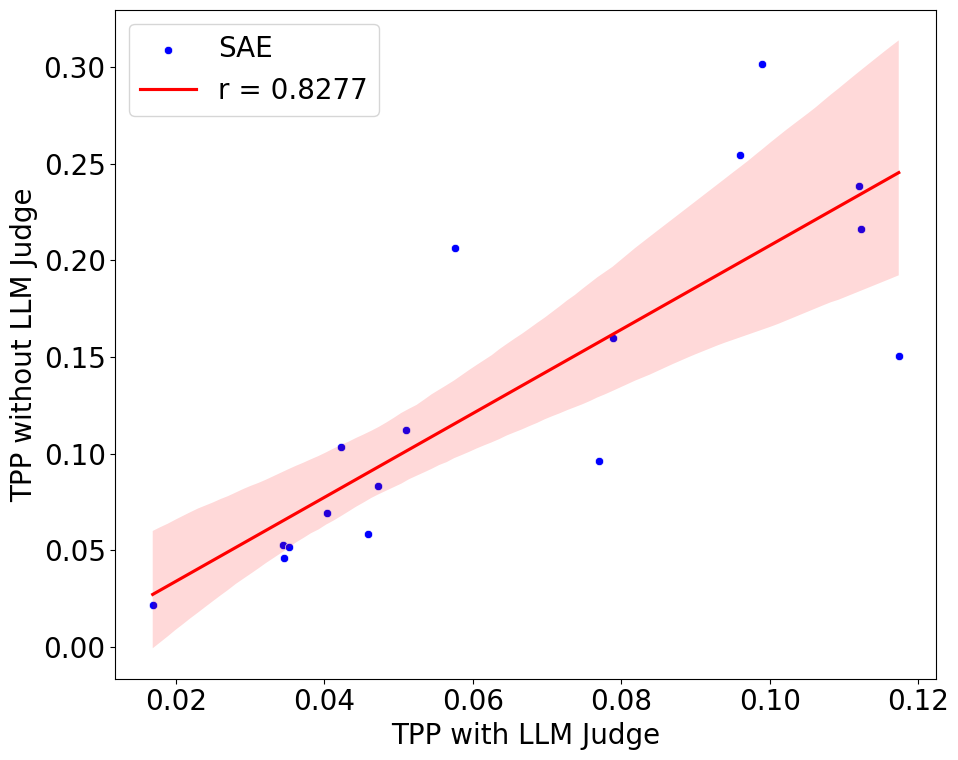}
    \caption{Correlation of TPP scores on Gemma-2-2B (left) and Pythia-70M before and after filtering SAE latents with the LLM judge described in Section \ref{sec:method:latent_selection}. The red area denotes the 95\% confidence interval.}
    \label{fig:tpp_corr}
\end{figure}

We measure the correlation of SHIFT (Figure \ref{fig:scr_corr} and TPP (Figure \ref{fig:tpp_corr}) scores with and without an interpretability filter as judged by an LLM. Incorporating an LLM judge provides an additional layer of validation by ensuring the identified latents are interpretable. The SHIFT and TPP metrics, when used without an LLM judge, offer simplicity, speed, and cost-effectiveness. These streamlined versions can be calculated rapidly, often within seconds, allowing for frequent assessments, such as during SAE training. This introduces a trade-off: while the LLM judge enhances confidence in the metric, it also increases complexity and computational requirements.

\clearpage
\subsection{LLM judge prompt}
\label{apx:prompt}

Our LLM judge prompt contains a system prompt, few-shot examples and a real example for labelling.

\textbf{System prompt}
\begin{lstlisting}
You are a meticulous AI researcher conducting an important investigation into a certain neuron in a language model. Your task is to analyze the neuron and score how strong its behavior is related to a concept in {concepts} on an integer scale from 0 to 4 in json format.

Task description:
You will be given a list of text examples on which the neuron activates. The specific tokens which cause the neuron to activate will appear between delimiters like <<this>>. The activation value of the token is given after each token in parentheses like <<this>>(3). 
You will also be shown a list called promoted tokens. The logits promoted by the neuron shed light on how the neuron's activation influences the model's predictions or outputs. It is possible that this list is more informative than the list of text examples.
For each concept, try to judge whether the neurons behavior is related to the concept.
If part of the text examples or predicted tokens are incorrectly formatted, please ignore them.
If you are not able to find any coherent description of the neurons behavior, decide that the neuron is not related to any concept.

Scoring rubric:
Score 4: The majority of examples, activation scores, and promoted tokens are clearly related to the concept.
Score 3: About half of the examples and promoted tokens are directly related to the concept. 
Score 2: Only some of the examples are directly related to the concept, and some more are distantly related.
Score 1: NONE of the examples is directly related to the concept, but single tokens can be distantly related to the general domain of the concept.
Score 0: NONE of the text examples can be distantly related in any way to the broader field of the concept.

Structure your response as follows:
Step 1. Give a single sentence summary for the full text examples.
Step 2. Give a separate single sentence summary for the promoted tokens.
Step 3. Discuss your decision in 1-3 sentences.
After finishing all steps above, provide a single json block at the end of your response. The json block should contain your scores on an integer scale from {min_scale} to {max_scale} for each concept as shown in the examples.

\end{lstlisting}

\textbf{Few-shot examples}
\begin{lstlisting}
Promoted tokens:  broadcasts, broadcasting, Broadcasting, television, broadcast, Television, announ,Television,TV, TV
Example prompts: 

Example 1: Radio Nova (Ireland)

Radio Nova was a pirate radio station <<broadcasting>>(2) from Dublin, Ireland. Owned and operated by the UK pirate radio veteran Chris Cary, the station's first broadcasts were during the summer of 1981 on 88.5 MHz FM and 819 kHz AM.

Early history
Prior to Nova's arrival, Irish radio consisted of the government broad<<caster>>(2) <<RT>>(2) and a number of local AM pirate <<stations>>(3). Radio Nova was the first <<station>>(2) in Ireland to use a high powered signal on FM. By 1982 Radio Nova was pulling in over 40% of the available audience around Dublin. In September 1982 Radio

Example 2: the network, and was the interim president of The Weather Channel for four months in 2013.

Scott is a 25-year veteran of NBC News. Before founding Peacock, she was executive producer and general manager of <<NBC>>(2) News Productions and NBC Media Productions. She was a member of the executive team for "Dateline" and "Now, with Tom Brokaw and Katie Couric."

Scott joined <<NBC>>(3) News in 1990 as news director for WTVJ-<<TV>>(2), <<NBC>>(3)'<<s>>(3) Owned and Operated station in Miami. Her honors include a number of national news Emmy awards in addition to a George Foster Pe

Chain of thought: Step 1: All activations are on words related to television and broadcasting.
Step 2: The top promoted logits are related to television and broadcasting.
Step 3: These themes are clearly related to filmmakers. I will rate filmmakers as a 4, and all other classes as 0.

{"gender": 0, "professor": 0, "nurse": 0, "accountant": 0, "architect": 0, "attorney": 0, "dentist": 0, "filmmaker": 4}
\end{lstlisting}

\textbf{Real example}
\begin{lstlisting}
"Okay, now here's the real task.
As a reminder, we only want to use these classes: Beauty_and_Personal_Care, Books, Automotive, Musical_Instruments, Software
Promoted tokens:  connector,  cable,  connectors,  connections,  cables
Example prompts: 

Example 1: Hager

The Hager Group is a leading supplier of solutions and services for electrotechnical installation in residential and commercial buildings as <<well>>(44) as for industrial applications.

As a leading supplier of systems, solutions and services for <<electrical>>(47) installations, the Hager Group provides an extensive offer, ranging from <<power>>(42) distribution and <<cable>>(103) management <<to>>(42) smart building automation and safety and security items - for an equally extensive field of application suitable for residential, commercial and industrial properties.

Besides the Hager brand which stands for a wide range of systems, solutions and electrotechnical components in buildings, the Hager Group is also home to the brands Daitem and Diagral offering security items

Example 2: the plaintiff and his family rather than the electrical company. As such our analysis of the facts focuses upon the information given to Spink when Floyd called to request service. Floyd merely indicated there was a "lack of power to his motor" and a "power shortage" was perhaps the cause. Floyd did not convey any information to Spink that the electrical problem could have originated between the meter box and the nonworking motor. Furthermore, since Floyd had not detected the <<hidden>>(34) short in <<the>>(51) <<extension>>(88) <<cord>>(46), a <<cord>>(39) he had exclusive control over, he could not have given sufficient information to raise this possibility to Spink. The two employees arrived at

Chain of thought:
\end{lstlisting}

\end{document}